\documentclass[letterpaper, 10 pt, conference]{ieeeconf}  

\IEEEoverridecommandlockouts                              

\usepackage{amsmath} 
\usepackage{amssymb}  
\usepackage{bm}
\usepackage[ruled,vlined,linesnumbered]{algorithm2e}
\usepackage{threeparttable} 
\usepackage{booktabs}       

\usepackage{multirow}       
\usepackage{rotating}       

\usepackage{graphicx}       

\usepackage{array}          

\usepackage{url}
\title{\LARGE \bf
I$^{2}$EKF-LO: A Dual-Iteration Extended
Kalman Filter Based \\ LiDAR Odometry
}

\author{Wenlu Yu$^{*}$,  Jie Xu$^{*}$,  Chengwei Zhao, Lijun Zhao, Thien-Minh Nguyen, Shenghai Yuan,\\ Mingming Bai, Lihua Xie,  \textit{Fellow, IEEE}
\thanks{*denotes that Wenlu Yu and Jie Xu contribute equally to this work.}
\thanks{This work was supported in part by the National Natural Science Foundation of China under Grand   62073101 and  Heilongjiang ``Hundreds and Thousands" Engineering Science and Technology Major Special Support Action Plan SC2021ZX02A0040.}
\thanks{Wenlu Yu, Jie Xu and Lijun Zhao are with State Key Laboratory of Robotics and Systems, Harbin Institute of Technology, Harbin 150001, China (e-mail: yuwenlu@stu.hit.edu.cn; jeff\_xu\_0503@foxmail.com; zhaolj@hit.edu.cn).}%
\thanks{Jie Xu, Thien-Minh Nguyen, Shenghai Yuan, Mingming Bai and Lihua Xie are with the School of Electrical and Electronic Engineering, Nanyang Technological University, Singapore 639798 (e-mail: jeff\_xu\_0503@foxmail.com; tmng@kth.se; shyuan@ntu.edu.sg; elhxie@ntu.edu.sg).}
\thanks{Chengwei Zhao is with Hangzhou Guochen Robot Technology Company Limited, Hangzhou 311200, China (e-mail: chengweizhao0427@gmail.com).}
\thanks{Mingming Bai is with State Key Laboratory of Industrial Control Technology, Zhejiang University, Hangzhou 310000, China (e-mail: mingmingbai@zju.edu.cn).}
\thanks{Lijun Zhao is the corresponding author.}
}%

\begin{document}

\maketitle

\begin{abstract}

LiDAR odometry is a pivotal technology in the fields of autonomous driving and autonomous mobile robotics. However, most of the current works focus on nonlinear optimization methods, and still existing many challenges in using the traditional Iterative Extended Kalman Filter (IEKF) framework to tackle the problem: IEKF only iterates over the observation equation, relying on a rough estimate of the initial state, which is insufficient to fully eliminate motion distortion in the input point cloud; the system process noise is difficult to be determined during state estimation of the complex motions; and the varying motion models across different sensor carriers. To address these issues, we propose the Dual-Iteration Extended Kalman Filter (I$^2$EKF) and the LiDAR odometry based on I$^2$EKF (I$^2$EKF-LO). This approach not only iterates over the observation equation but also leverages state updates to iteratively mitigate motion distortion in LiDAR point clouds. Moreover, it dynamically adjusts process noise based on the confidence level of prior predictions during state estimation and establishes motion models for different sensor carriers to achieve accurate and efficient state estimation. Comprehensive experiments demonstrate that I$^2$EKF-LO achieves outstanding levels of accuracy and computational efficiency in the realm of LiDAR odometry. Additionally, to foster community development, our code is open-sourced.\footnote{\url{https://github.com/YWL0720/I2EKF-LO}\label{foot_github}}
\end{abstract}

\section{INTRODUCTION}
With advancements in autonomous driving, augmented reality (AR), virtual reality (VR), and other technologies, Simultaneous Localization and Mapping (SLAM) has emerged as a prominent research area \cite{summarize}. LiDAR sensors (including mechanical rotating LiDAR and solid-state LiDAR) are favored for their ability to directly acquire environmental depth information, high measurement accuracy, and lack of susceptibility to external interference. They are often paired with Inertial Measurement Units (IMU) to estimate the motion state of the sensor carrier in real-time and construct point cloud maps of the environment. In the domain of LiDAR-Inertial odometry (LIO), FAST-LIO \cite{fast_lio, fast_lio2} represents a notable achievement, integrating IMU and LiDAR measurements closely through Iterative Extended Kalman Filter (IEKF) to achieve high precision and low time-overhead state estimation. Despite LIO systems achieving higher accuracy with additional IMU measurements, LiDAR odometry remains relevant in scenarios where: (1) the carrier's motion state exceeds the range of IMU, or the IMU fails \cite{point_lio}; (2) the carrier operates in a non-inertial reference frame \cite{underwater}; (3) during the calibration of LiDAR and IMU external parameters \cite{li_init, li_calib}.

\begin{figure}[t]
    \vspace{0.2cm}
	\centering 
	\includegraphics[width=0.47\textwidth]{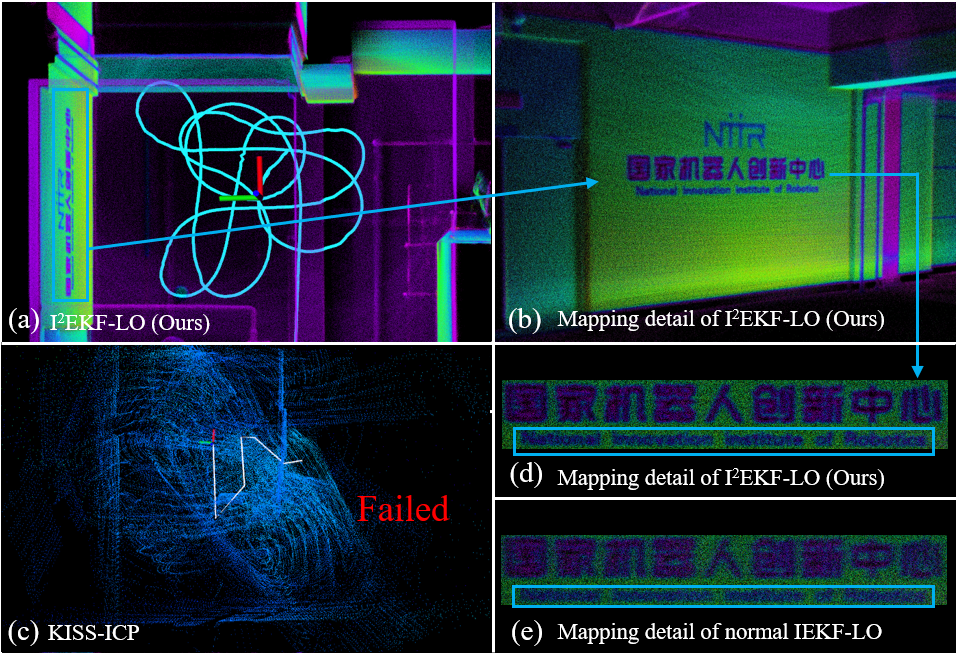}
	\caption{Mapping results of I$^2$EKF-LO in HIT-TIB dataset (sequence \textit{walk}). When I$^2$EKF-LO iterates only over the observation process, it degenerates to normal IEKF-LO. I$^2$EKF-LO has better handling of details compared to IEKF-LO. While KISS-ICP fails completely on this sequence using the same resolution.}
	\label{fig:tib}
\end{figure}

In recent years, there have been numerous excellent works in the field of LiDAR odometry, such as F-LOAM \cite{f_loam}, CT-ICP \cite{ct_icp}, KISS-ICP \cite{kiss_icp}, Traj-LO \cite{traj_lo}, MULLS \cite{mulls}, and DLO \cite{dlo},  etc., most of which are based on nonlinear optimization methods. Although the FAST-LIO series exemplifies the exceptional performance of Iterative Extended Kalman Filter in LIO systems, LiDAR odometry systems based on the Kalman filter framework are less common. FAST-LO \cite{li_init} inherits the philosophy of FAST-LIO under the IEKF framework, replacing the IMU with a uniform motion model to provide system priors. However, direct application of the IEKF framework in LiDAR odometry reveals several issues:

1) LO systems without additional sensor support often start with poor priors. The IEKF, on the other hand, only iterates the observation process and does not completely eliminate the motion distortion associated with the current state in the input point cloud.

2) In IEKF-based LIO systems, process noise is considered constant. However, the confidence in the uniform motion assumption for LO systems varies with changes in system motion state, leading to process noise fluctuation.

3) During state transitions, most existing methods treat rotation and translation separately. However, for a particular sensor carrier motion model, there may be a coupling between rotation and translation.

To address these challenges, we introduce I$^2$EKF-LO, a LiDAR odometry employing a dual Iterative Extended Kalman Filter. The term ``I$^2$" denotes the dual iteration process. It uses our I$^2$EKF designed for the specific problem of LiDAR odometry to precisely eliminate motion distortion in input point clouds. We also account for the differences in motion patterns across sensor carriers and dynamically adjust system process noise during state estimation. Specifically, our contributions are as follows:

1) We propose I$^2$EKF, a comprehensive dual Iterative Extended Kalman Filter framework for LiDAR odometry, iterating over both observation equations and the prediction process's distortion removal phase to improve point cloud quality and consequently, state estimation accuracy.

2) By leveraging measurement innovation \cite{akf}, we quantify the confidence level in the uniform motion model of the current state, dynamically adjusting process noise to enhance system robustness across various motion intensities.

3) Considering the coupling between rotation and translation based on different sensor carriers' motion models, we introduce SE(3) transformations to flexibly manage rotation and translation operations.

4) Extensive experimental testing on public datasets and in real-world environments demonstrates that I$^2$EKF-LO achieves outstanding levels of accuracy and computational efficiency among current LiDAR odometry algorithms. To further contribute to the community, our source code is made publicly available.
\section{RELATED WORKS}
LOAM \cite{loam}, is one of the classical representatives of LiDAR odometry. It divides the point cloud into two kinds: plane feature points and line feature points according to the spatial curvature of the point cloud, constructs constraints on the point-to-plane distance and the point-to-line distance, respectively, and solves for the state using the method of nonlinear optimization. LOAM operates in real-time at a frequency of 10 Hz, utilizing two point cloud registration modes: scan-to-scan and scan-to-map. However, its computational efficiency is limited due to the extensive processing required for thousands of point clouds and the absence of an effective map management strategy. F-LOAM is an improved version of LOAM, which omits the scan-to-scan process in favor of direct scan-to-map registration, significantly enhancing system computational efficiency while maintaining high accuracy. Nonetheless, F-LOAM does not account for motion distortion of the point clouds, compromising its performance in real-world settings. Lego-LOAM \cite{lego_loam}, another improvement on LOAM, boosts accuracy and operational efficiency through ground feature extraction and a two-step optimization process, making it more suitable for resource-constrained terrestrial robots. However, it does not adapt well to vehicles with non-horizontal LiDAR installations. Loam-Livox \cite{loam_livox}, tailored for solid-state LiDARs, addresses the challenges of narrow fields of view and irregular scanning patterns through quality feature extraction and dynamic target filtering, achieving improvements in accuracy and efficiency, though it becomes unstable with intense motion.

Dellenbach \textit{et al}. \cite{ct_icp} proposed CT-ICP, which introduced the concept of continuous time representation of sensor motion as a continuous function over time, estimating the motion state of the vehicle at the start and end of a point cloud frame. CT-ICP demonstrates exceptional accuracy on the Kitti dataset \cite{kitti}, but the increase in the dimensionality of the state leads to higher computational costs and convergence issues. KISS-ICP, a simple and efficient LiDAR odometry system, estimates states through point-to-point ICP without relying on extensive parameters, ensuring stable operation across various environments and motion patterns. However, its effectiveness diminishes with sparse point clouds. Yuan \textit{et al}. \cite{voxel_map} introduced VoxelMap, a LiDAR odometry based on a filtering framework that accounts for the uncertainty of LiDAR measurements and proposes an adaptive voxel map representation. It shows excellent precision on the Kitti dataset, focusing more on map feature representation and maintenance rather than addressing motion distortion in point clouds directly.

Our work is inspired by FAST-LIO \cite{fast_lio}, which constructs a discrete model of system motion, utilizes IMU integration for state prediction, and employs IMU measurements back propagation to accurately eliminate motion distortion in point clouds. By treating the distance from point to plane as the residual observation and using the IEKF (Iterative Extended Kalman Filter) for tight coupling of IMU and LiDAR measurements, it estimates the state. FAST-LIO2 \cite{fast_lio2} introduces an incremental kdtree \cite{ikdtree} for efficient and rapid map maintenance. I$^2$EKF-LO inherits the advantages of the FAST-LIO series, while being more suitable for working environments where IMU measurement information is missing and only point clouds input is available.

\section{OVERVIEW AND PREPROCESSING}
\subsection{System Overview} 

\begin{figure*}[t]
	\centering 
	\includegraphics[width=0.9\textwidth]{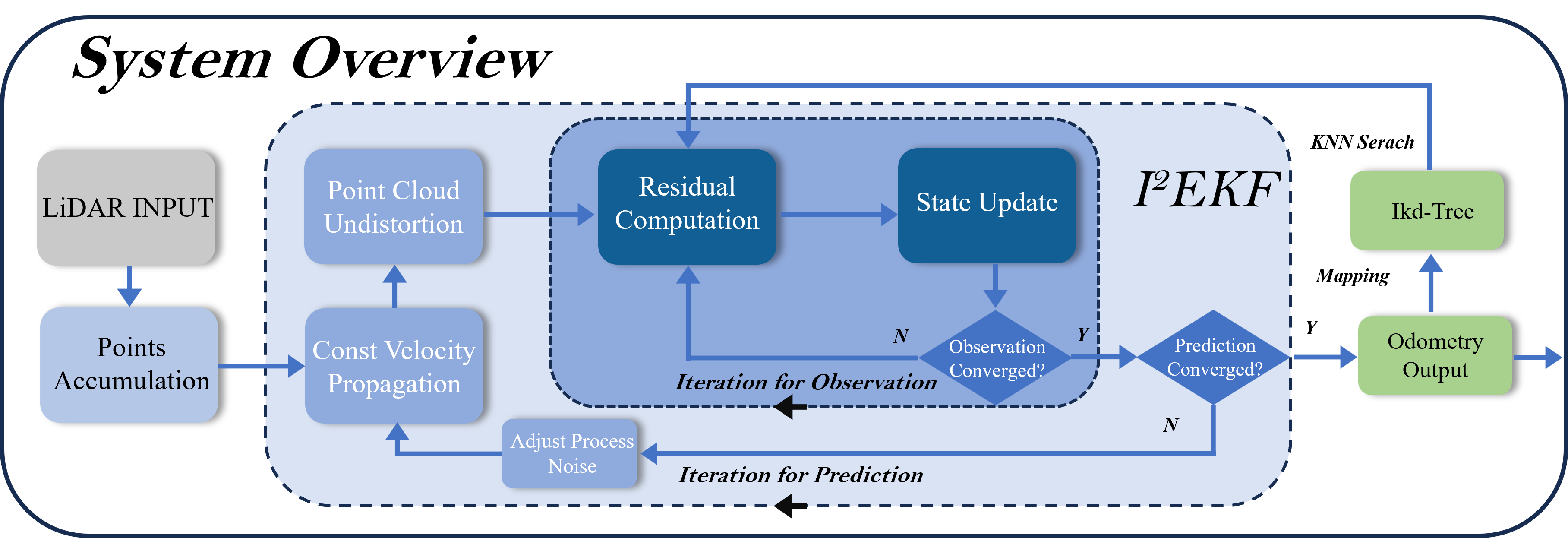}
	\caption{Framework of I$^{2}$EKF-LO.}
	\label{fig:overview}
\end{figure*}
The system architecture is illustrated in Figure. \ref{fig:overview}. I$^2$EKF-LO constructs a uniform motion model based on the type of the sensor carrier, serving as the predictive prior for the state of system. Depending on the intensity of the motion, it may be necessary to segment the incoming point cloud into frames. Utilizing I$^2$EKF, the system iteratively eliminates motion distortion within the point cloud, identifies point-to-plane matching relationships, and constructs point-to-plane distance residuals. Moreover, based on the confidence in the uniform motion assumption, it dynamically adjusts the process noise and finally integrates the prior to update the state. And the point cloud is registered to the world coordinate system using the posterior state, and an incremental kdtree \cite{ikdtree} is employed for management.

\subsection{Kinematic Model}
The state vector of system is defined as:
\begin{equation}
    \label{eq:state}
    \mathbf{x} \triangleq 
    \begin{bmatrix}
    \mathbf{R}^G & \mathbf{t}^G & \mathbf{v}^G & \bm{\omega}
    \end{bmatrix}^T \in \mathcal{M},
\end{equation}
where the state space $\mathcal{M} = SO(3) \times \mathbb{R}^{9}$, $G$ denotes the global coordinate system, and $\mathbf{R}^G, \mathbf{t}^G$ represent the rotation and translation of the LiDAR coordinate system in the global frame at the end of a point cloud frame, respectively. The linear velocity of the LiDAR in the global frame is denoted by $\mathbf{v}^G$, and its angular velocity in the body frame is represented by $\bm{\omega}$. These velocities are modeled as random walks driven by Gaussian noise $\mathbf{n}_v, \mathbf{n}_\omega$.
\subsubsection{Uniform Motion Model}
The discrete state transition is formulated based on the uniform motion model as follows:
\begin{equation}
\mathbf{x}_{k+1} = \mathbf{x}_k \boxplus ( \mathbf{f}(\mathbf{x}_k, \mathbf{w}_k)),
\end{equation}
here, $\boxplus$ denotes the ``plus" on the state manifold defined in \cite{fast_lio, fast_lio2}, and $\mathbf{w}_k = [\mathbf{n}_{v, k} \ \ \mathbf{n}_{\omega, k}]^T$ represents the process noise.
\begin{itemize}
    \item \textbf{Uniform Motion Model 1}: For handheld devices or certain aerial vehicles (where rotation and translation are weakly coupled), the uniform motion assumption is as follows: during the duration $t_k$ of the $k^{th}$ LiDAR point cloud frame, both the linear velocity in the global frame and the angular velocity in the body frame are assumed constant and equal to those in the duration of the $(k-1)^{th}$ frame. And the state transition function is defined as
    \begin{equation}
    \label{eq:assumption_a}
        \mathbf{f}(\mathbf{x}, \mathbf{w}) = 
    \begin{bmatrix}
        \bm{\omega} \Delta t\\
        \mathbf{v}^G \Delta t\\
        \mathbf{n}_v \Delta t\\
        \mathbf{n}_\omega \Delta t\\
    \end{bmatrix}.
    \end{equation}
    
    \item \textbf{Uniform Motion Model 2}: For ground-moving robots (where rotation and translation are strongly coupled), the uniform motion assumption is defined as: during the duration $t_k$ of the $k^{th}$ LiDAR point cloud frame, both the linear velocity and the angular velocity in the body frame are assumed constant and equal to those in the duration of the $(k-1)^{th}$ frame. And the state transition function is defined as
    \begin{equation}
    \label{eq:assumption_b}
        \mathbf{f}(\mathbf{x}, \mathbf{w}) = 
    \begin{bmatrix}
        \bm{\omega} \Delta t\\
        (\mathbf{R}^G \boxplus \bm{\omega} \Delta t) (\mathbf{R}^G)^{-1} \mathbf{v}^G \Delta t\\
        ((\mathbf{R}^G \boxplus \bm{\omega} \Delta t) (\mathbf{R}^G)^{-1} - \mathbf{I}) \mathbf{v}^G +  \mathbf{n}_v \Delta t \\
        \mathbf{n}_\omega \Delta t\\
    \end{bmatrix}.
    \end{equation}
    
\end{itemize}

The temporal interval between two consecutive LiDAR point cloud frames is denoted by $\Delta t$, facilitating the definition of the system's dynamical evolution over time.
\subsubsection{Observation Model}
Consistent with the approach outlined in \cite{fast_lio2}, the proposed I$^2$EKF-LO framework abstains from extracting features from the raw point clouds. Instead, it constructs an observation model based on the direct calculation of point-to-plane distances. This methodology leverages the detailed information inherent in the point clouds to enhance the accuracy of the state estimation. The observational relationship is formalized as follows:

\begin{equation}
\label{eq:meas}
0 = \mathbf{h}_j(\mathbf{x}_k, \mathbf{n}_j^L) \triangleq \mathbf{u}_j^T(\mathbf{R}_k^G (\mathbf{p}_j^L + \mathbf{n}_j^L) + \mathbf{t}^G_k - \mathbf{q}_j^G),
\end{equation}
where $\mathbf{n}_j^L$ represents the LiDAR measurement noise, and $\mathbf{p}_j^L$ denotes the coordinates of a point in the LiDAR frame. The vector $\mathbf{u}_j^T$ is the normal vector of the plane matched to the point $\mathbf{p}_j^L$ within the map, while $\mathbf{q}_j^G$ is a point on this plane.
\subsection{Point Cloud Preprocessing}
The accuracy of the distortion removal process in solid-state LiDAR, characterized by its non-repetitive scanning pattern, is notably influenced by the initial state estimation. Inspired by \cite{loam_livox}, the I$^2$EKF-LO framework incorporates a preprocessing stage where the point cloud of the current frame, spanning a time interval $\Delta t$, is subdivided temporally into $n$ segments. This segmentation reduces the time interval for each segment to $\frac{1}{n} \Delta t$. Such a preprocessing strategy narrows the gap between the initial state provided by the uniform motion assumption and the true state, thereby facilitating more accurate results, particularly in scenarios involving rapid vehicular movements.

\section{DUAL-ITERATION EXTENDED \\ KALMAN FILTER}
\subsection{IEKF Review}
Let's recall why the IEKF framework of the FAST-LIO series is solved iteratively? Firstly, the quality of point cloud matching depends on the accuracy of the current frame's state estimation. Transforming the current point cloud into the world coordinate system using a state that is closer to the true state results in more accurate matching. This improved matching, in turn, contributes to deriving a more refined solution. Furthermore, due to the nonlinearity of the observation equation, linearizing it with a state approximation closer to the true state helps minimize the errors introduced during the linearization process. This enhanced degree of linearization leads to better solutions. This kind of ``chicken-laying-egg, egg-laying-chicken" relationship needs to be solved iteratively to obtain the optimal value. The iteration in IEKF is specifically concentrated around the observation component. In LiDAR odometry problem, without the assistance of IMU measurements, the prediction of the initial value is of poor quality, which will bring a large error disturbance to the input point cloud during the undistortion process. Similarly, the process of point cloud undistortion also follows the aforementioned cyclical logic. Only by using a state that closely approximates the true state to undistort the point cloud can achieve more accurate point cloud data, which, in turn, facilitates a better estimation in the subsequent iterations.

Hence, we introduce the Dual-Iteration Extended Kalman Filter (I$^2$EKF), which performs additional iterations on the distortion removal step of prediction process to minimize the impact of point cloud distortion, while concurrently iterating on the observation process to mitigate mismatches and nonlinear effects. Moreover, recognizing the varying confidence in the uniform motion model depending on the vehicular motion state, the I$^2$EKF dynamically adjusts the process noise within its iterations to enhance the system's robustness across different motion conditions.
\subsection{Forward Propagation in I$^2$EKF Uniform Motion Model}

Similar to \cite{fast_lio}, I$^2$EKF-LO adopts the following forward propagation relationship:
\begin{equation}
\label{eq:forword}
    \begin{aligned}
        \widehat{\mathbf{x}}_{k+1} &= \bar{\mathbf{x}}_{k} \boxplus  \mathbf{f}(\bar{\mathbf{x}}_{k}, \mathbf{0}), \\
        \widehat{\mathbf P}_{k+1} &= \mathbf F_{\widetilde{\mathbf x}_k}\bar{\mathbf P}_{k}\mathbf F_{\widetilde{\mathbf x}_k}^T + \mathbf F_{\mathbf w_k} \mathbf Q_k \mathbf F_{\mathbf w_k}^T,
    \end{aligned}
\end{equation}
where $\bar{\mathbf{x}}_{k}$ and $\bar{\mathbf P}_{k}$ represent the posterior state and covariance at frame $k$, respectively. Similarly, $\widehat{\mathbf{x}}_{k+1}$ and $\widehat{\mathbf P}_{k+1}$ denote the predicted state and covariance for frame $k+1$. The matrix $\mathbf Q_k$ is defined as the covariance of the noise $\mathbf{w}_k$. The matrices $\mathbf F_{\widetilde{\mathbf x}_k}$ and $\mathbf F_{\mathbf w_k}$ are specified as follows:
\begin{equation}
\begin{split}~\label{eq:f_x_w}
\mathbf F_{\widetilde{\mathbf x}_k} &= \left. \textstyle \frac{\partial \left({\mathbf x}_{k+1} \boxminus \widehat{\mathbf x}_{k+1}\right)}{\partial \widetilde{\mathbf x}_k}\right|_{\begin{subarray}{l}\widetilde{\mathbf x}_k = \mathbf 0 ,\ \mathbf w_k = \mathbf 0 \end{subarray}}, \\
\mathbf F_{\mathbf w_k} &=\left. \textstyle \frac{\partial \left({\mathbf x}_{k+1} \boxminus \widehat{\mathbf x}_{k+1}\right)}{\partial \mathbf w_k}\right|_{\begin{subarray}{l}\widetilde{\mathbf x}_k = \mathbf 0 ,\ \mathbf w_k = \mathbf 0 \end{subarray}}, \\
\end{split}
\end{equation}
where the error state vector is denoted by $\widetilde{\mathbf x}_k = \mathbf x_k \boxminus \widehat{\mathbf x}_k$ and $\boxminus$ denotes the ``minus" on the state manifold which defined in \cite{fast_lio, fast_lio2}.

\subsection{Iterative Update in I$^2$EKF}

Each iterative update encompasses two primary components: the iterative distortion correction update and the iterative observation update. Additionally, based on the results of the first convergence within observation iteration, we dynamically adjust the process noise.

\subsubsection{Iterative Distortion Correction Update}
Given that the LiDAR sensor is not stationary during the collection of each point cloud frame, the raw point clouds contain motion-induced distortions. To mitigate these distortions, I$^2$EKF transforms the point cloud of the current frame \(k\), acquired at time \(\Delta t_k\), to the LiDAR coordinate system at the end of frame \(k\) at time \(t_{k}\). For a point \(p_i^{L_i}\) in the LiDAR coordinate system generated at time \(t_i\) within frame \(k\), given the predicted state transformation \(\widehat{\mathbf{T}}_k^G\) for frame \(k\) and the posterior state transformation \(\bar{\mathbf{T}}_{k-1}^G\) for frame \(k-1\), I$^2$EKF seeks the transformation \(\widehat{\mathbf{T}}_i^G\) of the LiDAR coordinate system in the world coordinate system at time \(t_i\), thereby obtaining the distortion-corrected point.

\begin{equation}
\label{eq:distort update}
    p_i^{L_{k}} = (\widehat{\mathbf{T}}^G_{k})^{-1} \widehat{\mathbf{T}}^G_i p_i^{L_i},    
\end{equation}
where
\begin{equation}
    \mathbf{T}^G = 
    \begin{bmatrix}
        \mathbf{R}^G | \mathbf{t}^G
    \end{bmatrix}.
\end{equation}

Regarding the two uniform motion models mentioned in Section \uppercase\expandafter{\romannumeral3}-B, the respective formulas are as follows:
\begin{itemize}
    \item \textbf{For uniform motion model 1}
    \begin{equation}
    \label{eq:undistort1}
    \begin{aligned}
        \widehat{\mathbf{R}}_i^G &= \bar{\mathbf{R}}_{k-1}^G . \text{interpolate}(\text{scale},  \mathbf{R}_{k}^G)\\
        \widehat{\mathbf{t}}_i^G &= \bar{\mathbf{t}}_{k-1}^G + \text{scale} (\mathbf{t}_{k}^G-\mathbf{t}_{k-1}^G)\\
        \text{scale} &= \frac{t_i - t_{k-1}}{t_k - t_{k-1}}
    \end{aligned}
    \end{equation}
    \item \textbf{For uniform motion model 2}
    \begin{equation}
    \label{eq:undistort2}
    \begin{aligned}
        \widehat{\mathbf{T}}_i^G &= \bar{\mathbf{T}}_{k-1}^G . \text{interpolate}(\text{scale},  \widehat{\mathbf{T}}_{k}^G)\\
        \text{scale} &= \frac{t_i - t_{k-1}}{t_k - t_{k-1}}
    \end{aligned}
    \end{equation}
\end{itemize}

The above descriptions correspond to the linear interpolation of rotation matrices and translation vectors independently, and the coupled linear interpolation within SE(3), to reflect the interrelation between rotation and translation across various motion patterns.

\subsubsection{Iterative Observation Update}
The observation equation is expanded to first order at the state of the \(\kappa^{th}\) observation iteration. For the $j^{th}$ point in frame $k$, the observation equation is as follows

\begin{equation}\label{eq:meas_jocob}
\begin{split}
    \mathbf 0 &= \mathbf h_j \left( \mathbf x_k, {\mathbf n}_{j}^L \right) \simeq \mathbf h_j \left( \widehat{\mathbf x}_k^\kappa, \mathbf 0 \right) + \mathbf H_j^\kappa  \widetilde{\mathbf x}_{k}^\kappa + \mathbf v_j\\&=  \mathbf z_j^\kappa + \mathbf H_j^\kappa  \widetilde{\mathbf x}_{k}^\kappa + \mathbf v_j.
\end{split}
\end{equation}

Within this context, \(\mathbf{H}_j^\kappa\) represents the Jacobian matrix of \(\mathbf{h}_j(\widehat{\mathbf{x}}_k^\kappa \boxplus \widetilde{\mathbf{x}}_k^\kappa, {\mathbf{n}}_{j}^L)\) evaluated at \(\widetilde{\mathbf{x}}_k^\kappa = \mathbf{0}\), where \(\mathbf{v}_j\) is drawn from the original measurement noise \(\mathbf{n}_{j}^L\) with a distribution \(\mathcal{N}(\mathbf{0}, \mathbf{R}_j)\). Utilizing the latest point cloud matching relationships, the observation residuals and observation matrices are calculated according to (\ref{eq:meas}). By integrating the uniform motion prior, the following maximum a-posteriori estimation (MAP) problem is formulated:
\begin{equation}
\begin{split}~\label{eq:error_states_solution}
    \min_{\widetilde{\mathbf x}_{k}^\kappa} \left( \| \mathbf x_k \boxminus \widehat{\mathbf x}_k \|^2_{ \widehat{\mathbf P}^{-1}_k} + \sum\nolimits_{j=1}^{m} \| \mathbf z_j^\kappa + \mathbf H_j^\kappa \widetilde{\mathbf x}_{k}^\kappa \|^2_{\mathbf R^{-1}_j} \right),
\end{split}
\end{equation}
let $\mathbf H\! = \![ \mathbf H_1^{\kappa^T}, \cdots, \mathbf H_m^{\kappa^T}]^T$, $\mathbf R\! =\! \text{diag}\left(\mathbf R_1, \cdots \mathbf R_m \right)$,$\mathbf P\! =\! \widehat{\mathbf P}_{k}$, and $\mathbf z_k^\kappa = \left[ \mathbf z_1^{\kappa^T}, \cdots, \mathbf z_m^{\kappa^T} \right]^T$.

The formula for updating the state is as follows
\begin{equation}
\begin{split}~\label{eq:kalman_gain}
   \widehat{\mathbf x}_{k}^{\kappa+1} &\! \!  = \!  \widehat{\mathbf x}_{k}^{\kappa} \! \boxplus \!  \left( -\mathbf K  {\mathbf z}_k^\kappa  - (\mathbf I - \mathbf K \mathbf H ) \left( \widehat{\mathbf x}_{k}^{\kappa} \boxminus \widehat{\mathbf x}_{k} \right)  \right),
    \end{split}
\end{equation}
where
\begin{equation}
\begin{split}~\label{eq:kalman_gain_new}
    \mathbf K \!\!=\!\! \left(\mathbf H^T {\mathbf R}^{-1} \mathbf H \!+\! {\mathbf P}^{-1} \right)^{-1}\!\mathbf H^T \mathbf R^{-1}.
    \end{split}
\end{equation}
\subsubsection{Process Noise Update}

In the actual motion of the carrier, the confidence in the constant velocity assumption changes over time. Unlike the general IEKF framework, which considers process noise as a constant, the I$^2$EKF dynamically corrects the process noise. Inspired by \cite{akf}, the relationship for the filter's process noise is as follows:
\begin{equation}
    \mathbf{w}_{k} = \mathbf{K} \mathbf{d}_{k},
\end{equation}
where \(\mathbf{d}_{k}\) is the innovation sequence \cite{akf}, and \(\mathbf{K}\) is the Kalman gain. Consequently,
\begin{equation}
    \mathbf{Q}_{k} = E[\widehat{\mathbf{w}}_{k}\widehat{\mathbf{w}}_{k}^T] = \mathbf{K} E[\mathbf{d}_k \mathbf{d}^T_k] \mathbf{K}^T.
\end{equation}

Thus, utilizing the innovation sequence obtained at the first convergence of the observation iteration, we quantify the discrepancy between the observation of point cloud point-plane residuals and the prediction of the constant velocity assumption, reflecting the current confidence in the constant velocity assumption. The process noise covariances for the velocity and angular velocity in the state vector are defined as \(\mathbf{Q}_{v}\) and \(\mathbf{Q}_{\omega}\), respectively.

\begin{equation}
\begin{aligned}
    \mathbf{Q}_{v} &= 
    cov_{v}^{scale} \Delta t^2 \mathbf{I}_{3\times 3}  \\ 
    \mathbf{Q}_{\omega} &= 
    cov_{\omega}^{scale} \Delta t^2 \mathbf{I}_{3\times 3}  
\end{aligned}
\end{equation}

Based on our experience from a large number of tests, the appropriate value of $cov_{v}^{scale}$ and $cov_{\omega}^{scale}$ is between 0.01 and 100. The larger value of $cov_{v}^{scale}$ and $cov_{\omega}^{scale}$, the less sensitive it is to the violent movement of the sensor. Therefore, we establish the following mapping between the innovation sequence and the process noise covariance parameters:
\begin{equation}
\label{e:dynamic_q}
    \mathbf{Q}_k = (\frac{100}{1 + e^{(-\alpha ||\mathbf{K}\mathbf{d}_k|| + \gamma)}} + 0.01)  \Delta t^2 \mathbf{I}_{3\times 3},
\end{equation}
where \(\alpha\) and \(\gamma\) are hyperparameters based on the movement scale. The process noise, corrected internally through the variation of the innovation sequence, adapts to the different intensities of motion.

The complete state estimation process of the I$^2$EKF is outlined as Algorithm \ref{alg:iekf}.

\begin{algorithm}
    \caption{{I$^2$EKF State Estimation}}
    \label{alg:iekf}
    \SetKwInOut{Input}{Input}\SetKwInOut{Output}{Output}\SetKwInOut{Start}{Start}\SetKwInOut{blank}{}
    \Input{Last optimal estimation $\bar{\mathbf x}_{k-1}$ and $\bar{\mathbf P}_{k-1}$, \\
    LiDAR  points ${\mathbf p}_{j}^{L_j}$ in current scan.}
    Forward propagation to obtain state prediction $\widehat{\mathbf x}_k$ via (\ref{eq:forword}) and covariance prediction $\widehat{\mathbf P}_k$ via (\ref{eq:f_x_w});\\
    $\alpha = -1$, $\kappa = -1$, $\widehat{\mathbf x}_{k}^{\alpha = 0} = \widehat{\mathbf x}_{k}$ $\widehat{\mathbf x}_{k}^{\kappa = 0} = \widehat{\mathbf x}_{k}$; \\
        \Repeat{$\| \widehat{\mathbf x}_{k}^{\alpha+1} \boxminus \widehat{\mathbf x}_{k}^\alpha\|< \epsilon_1$}{
        $\alpha = \alpha + 1$\\
        \textbf{Iterative undistort}\\  Perform distortion correction on the point cloud to obtain ${\mathbf p}_{j}^{L_{k}}$ via (\ref{eq:undistort1}) or (\ref{eq:undistort2}); \\
        \textbf{Process Noise Update}\\
        \If {$\alpha = 1$} { Adjust $\mathbf{Q}_k$ and repropagation  via (\ref{e:dynamic_q}) and (\ref{eq:forword});}
        \Repeat{$\| \widehat{\mathbf x}_{k}^{\kappa+1} \boxminus \widehat{\mathbf x}_{k}^\kappa\|< \epsilon_2$}{
        $\kappa =\kappa+1$;\\
        \textbf{Iterative measurement}\\  Compute residual $\mathbf z_j^{\kappa}$  and Jocobin $\mathbf H_j^\kappa$ via (\ref{eq:meas_jocob});\\
        Compute the state update $\widehat{\mathbf x}_{k}^{\kappa+1}$ via (\ref{eq:kalman_gain}) with the Kalman gain $\mathbf K$ from (\ref{eq:kalman_gain_new});\\ }
        $\widehat{\mathbf x}_{k}^{\alpha + 1} = \widehat{\mathbf x}_{k}^{\kappa+1}$;\\}
    $\bar{\mathbf x}_{k} = \widehat{\mathbf x}_{k}^{\alpha+1}$; $\bar{\mathbf P}_{k} = \left( \mathbf I - \mathbf K \mathbf H \right) {\mathbf P}$.\\
    \Output{Current optimal estimation $\bar{\mathbf x}_{k}$ and $\bar{\mathbf P}_{k}$.}
\end{algorithm}

\section{EXPERIMENTS}

To demonstrate the superior performance of I$^2$EKF-LO, extensive experiments were conducted in both public datasets and real-world environments. The results indicate that I$^2$EKF-LO achieves superior accuracy and real-time performance compared to several other prominent LiDAR odometry algorithms. All experiments were conducted on the ROS operating system, utilizing an Intel i7-12700H CPU with 16GB of RAM, and all algorithms were tested under identical parameters.

\subsection{Datasets}

\subsubsection{NTU VIRAL Dataset \cite{ntu}} The NTU VIRAL dataset is a multi-sensor dataset of outdoor scenarios for autonomous unmanned aerial vehicles (UAVs), whose ground truth was obtained via the Leica Nova MS60, and contains a variety of challenging scenarios. Given the relatively poor stability of drone movements, this dataset presents numerous difficulties for LiDAR odometry. We tested using the \textit{eee} and \textit{rtp} sequences from this dataset, employing a 16-beam Ouster LiDAR as the data source.

\subsubsection{M2DGR Dataset \cite{M2DGR}}M2DGR is a multisensor dataset for ground mobile robots. We utilized the \textit{room} sequence collected in indoor environments, with ground truth acquired via a motion capture system, and a Velodyne 32-beam LiDAR as the data source.

\subsubsection{Urbanloco Dataset \cite{urbanloco}}The Urbanloco dataset caters to urban environments for autonomous driving vehicles, with a Velodyne 32-beam LiDAR as the data source and ground truth obtained through high-precision GPS-RTK.

\subsubsection{HIT-TIB Dataset}
\begin{figure}[t]
	\centering 
	\includegraphics[width=0.47 \textwidth]{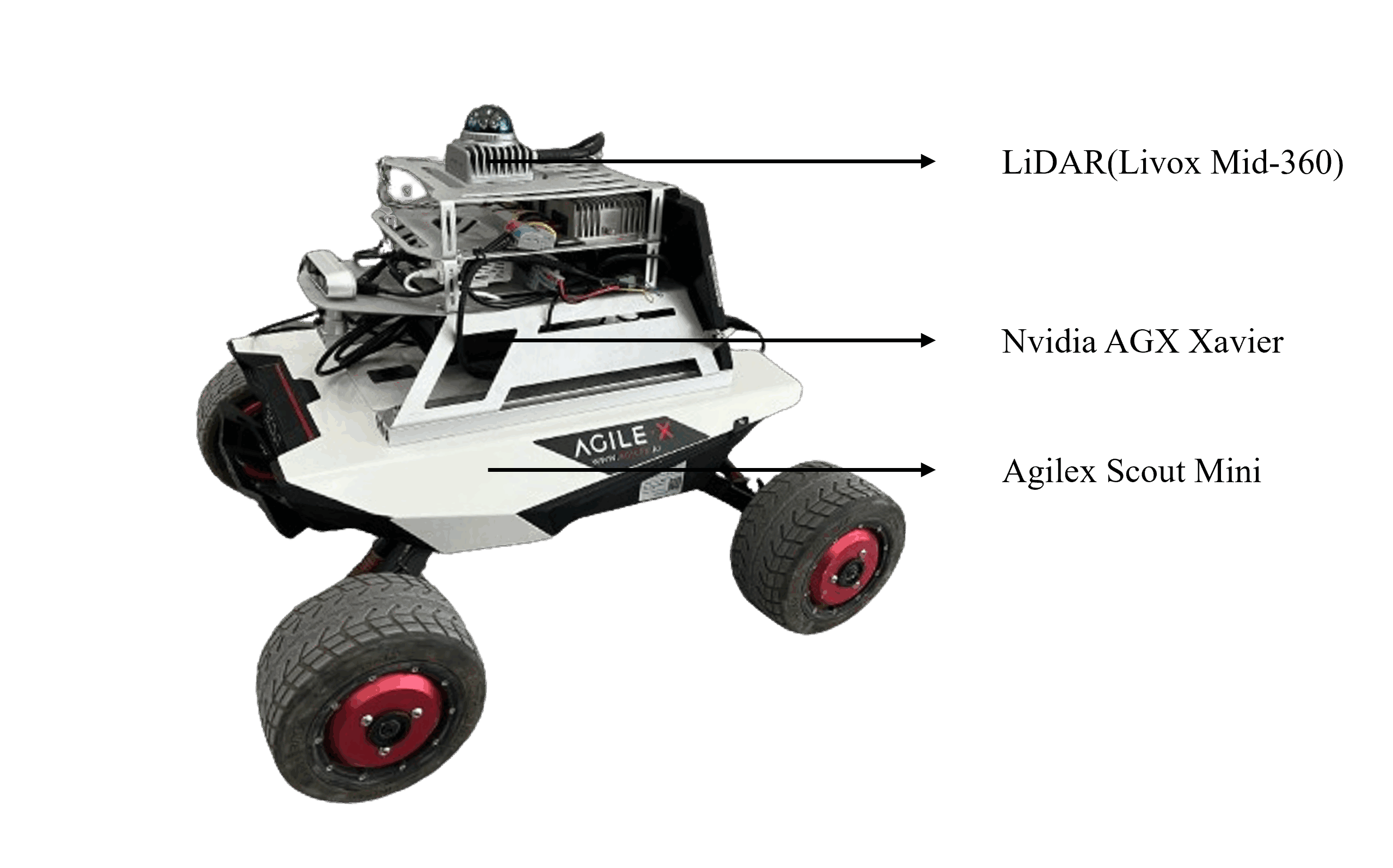}
	\caption{The experimental platform uses the Agilex Scout Mini as the mobile chassis, featuring four-wheel differential steering. It is equipped with a Livox Mid-360 LiDAR and uses the Nvidia AGX Xavier as the computing platform.}
	\label{fig:robot}
\end{figure}

The HIT-TIB dataset comprises solid-state LiDAR data collected in indoor scenarios by ground mobile robots, as illustrated in Fig. \ref{fig:robot}, including sequences from parking lot, hall, and long corridor.  

 \begin{table*}[t]
\footnotesize
\centering
\caption{Absolute translational errors (RMSE, meters) in Public Datasets}
\label{tab:rmse_benchmark}
\begin{threeparttable}
\begin{tabular}{@{}lcccccccccc@{}}
\toprule
 & \textit{eee\_1} & \textit{eee\_2} & \textit{eee\_3} & \textit{rtp\_1} & \textit{rtp\_2} & \textit{rtp\_3} & \textit{room\_1} & \textit{room\_2} & \textit{room\_3} & \textit{urbanloco\_1}  \\ \midrule
F-LOAM  & 10.492       & 9.606        & 1.024        & 11.108       & 6.584        & 8.590        & 0.168        & 0.146        & 0.223 & 3.632          \\
KISS-ICP        & 3.081        & 1.755        & 1.508        & 9.551        & 6.140        & 4.110        & 0.213        & 0.406        & 0.227 & 3.004             \\
CT-ICP         & 0.296        & 0.252       & ×        & ×             & \textbf{0.478} & 0.488      & 0.160        & 0.129        & 0.166 & 1.162                \\
I$^2$EKF-LO(w/o i2) & \textbf{0.232}& 0.201        & 0.237        & 6.026        & \underline{0.491}        & 0.831       & \textbf{0.150}        & \underline{0.127}        & 0.191    & 1.000             \\
I$^2$EKF-LO(M 1)        & \underline{0.290}        & \textbf{0.194}& \textbf{0.229}& \textbf{3.653}& 0.508     & \underline{0.368} & \underline{0.151}& \textbf{0.122}& \underline{0.162}& \underline{0.999}\\
I$^2$EKF-LO(M 2)        & 0.294        & \underline{0.198}& \underline{0.231}& \underline{3.701}& 0.512     & \textbf{0.360} & \textbf{0.150}& \textbf{0.122}& \textbf{0.161}& \textbf{0.987}\\
\bottomrule
\end{tabular}
\begin{tablenotes}
\footnotesize
\item[1] `×' denotes that the system totally failed. The best results overall are in \textbf{blod}, while the second best results are \underline{underlined}.
\item[2] ``M 1" denotes that the system uses uniform motion model 1 in (\ref{eq:assumption_a}). ``M 2" denotes that the system uses uniform motion model 2 in (\ref{eq:assumption_b}).
\end{tablenotes}
\end{threeparttable}
\end{table*}

 \begin{table}[h]
\footnotesize
\centering
\caption{End to end errors (meters) in HIT-TIB Datasets}
\label{tab:hit-ee}
\begin{threeparttable}
\begin{tabular}{@{}lccc@{}}
\toprule
 & \textit{hall} & \textit{parking} & \textit{corridor}   \\ \midrule
KISS-ICP        & 0.728       & 0.332        & 0.060($<$ 0.1)       \\
I$^2$EKF-LO(w/o i2) &      0.053($<$ 0.1)     &          0.074($<$ 0.1)     &      0.075($<$ 0.1)     \\
I$^2$EKF-LO        & \textbf{0.050($<$ 0.1)}        & \textbf{0.054($<$ 0.1)}& \textbf{0.058($<$ 0.1)}\\
\bottomrule
\end{tabular}
\begin{tablenotes}
\footnotesize
\item[1] The best results overall are in \textbf{blod}.
\end{tablenotes}
\end{threeparttable}
\end{table}

 \begin{table}[h]
\footnotesize
\centering
\caption{The Comparsion of average time consumption perscan (ms)}
\label{tab:time}
\begin{threeparttable}
\begin{tabular}{@{}lcccc@{}}
\toprule
 & \textit{room\_1} &\textit{room\_2} & \textit{room\_3} & \textit{urbanloco\_1}   \\ \midrule
 F-LOAM&  19.07&         19.21&          21.13&      38.98\\
KISS-ICP &  18.90&         \textbf{15.34}&          26.48&      23.76\\
CT-ICP &  35.31&         35.09&          33.31&      24.39\\
I$^2$EKF-LO &  \textbf{15.76}&         18.20&          \textbf{17.78}&      \textbf{23.62}\\
\bottomrule
\end{tabular}
\begin{tablenotes}
\footnotesize
\item[1] The best results overall are in \textbf{blod}.
\end{tablenotes}
\end{threeparttable}
\end{table}

\begin{figure}[h]
	\centering 
	\includegraphics[width=0.47\textwidth]{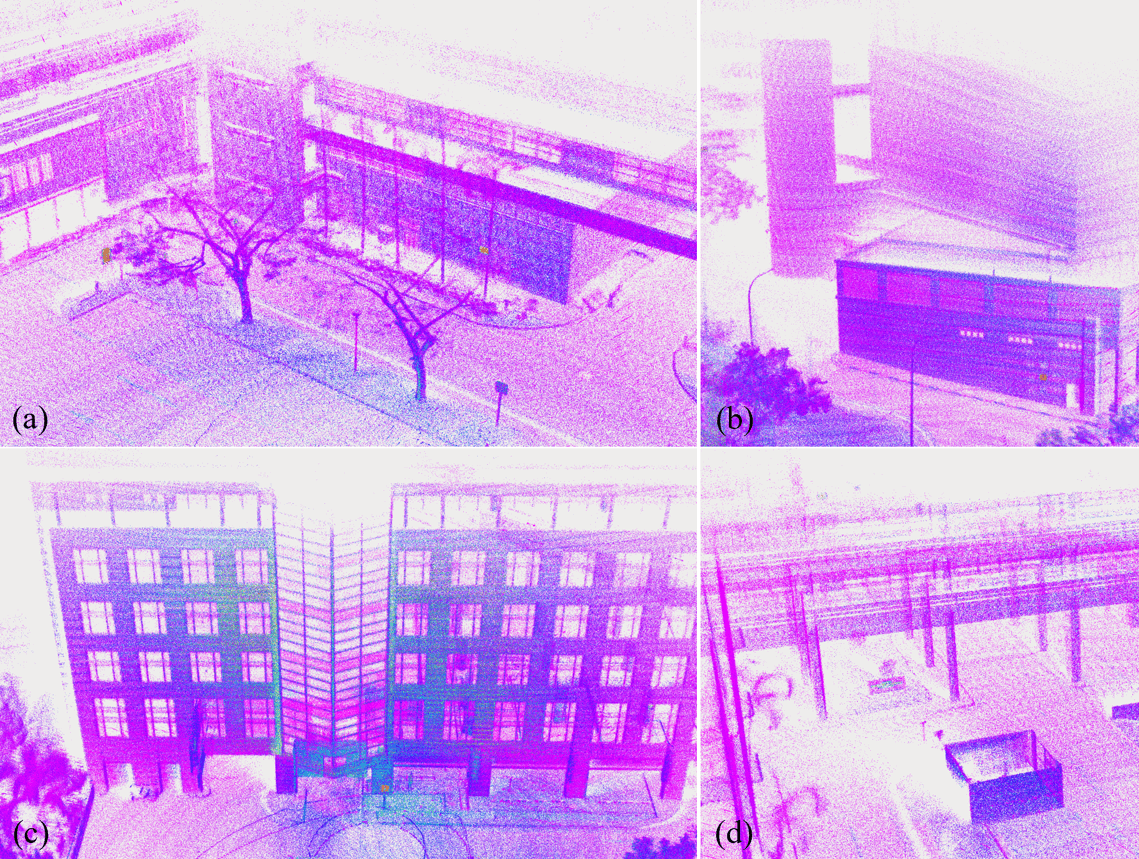}
	\caption{Mapping results of I$^2$EKF-LO in NTU VIRAL dataset.}
	\label{fig:NTU_res}
\end{figure}

\begin{figure}[h]
	\centering 
	\includegraphics[width=0.47\textwidth]{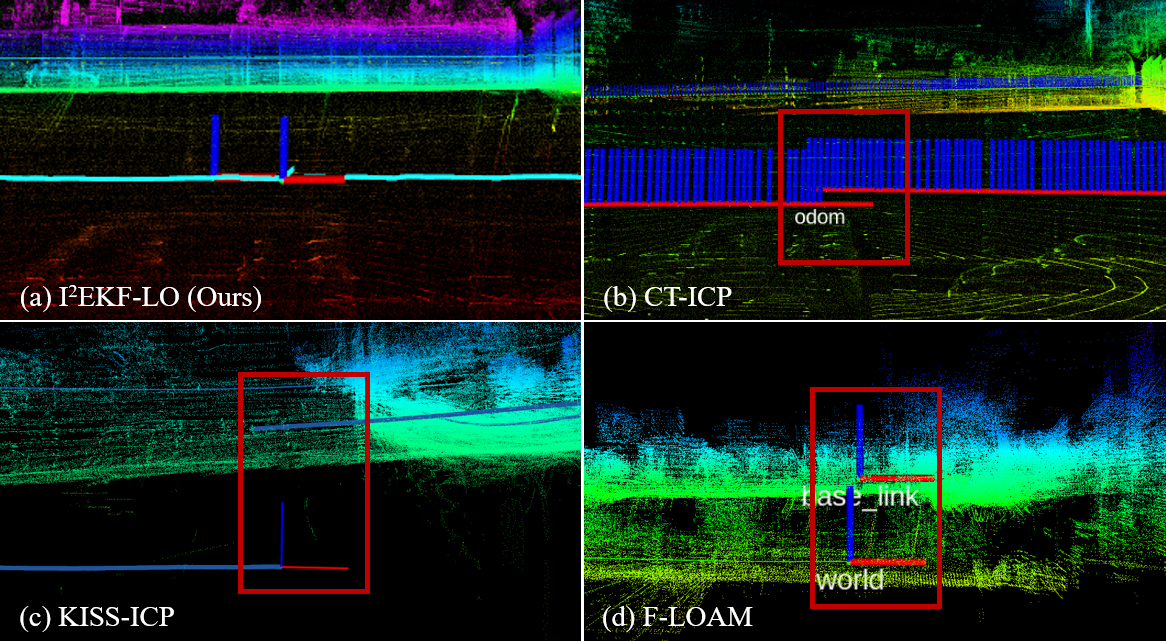}
	\caption{Mapping results in Urbanloco dataset. Significant $z$-axis errors in (b) CT-ICP, (c) KISS-ICP, and (d) F-LOAM as the lidar returns to the vicinity of the origin. While (a) I$^2$EKF-LO has a much smaller error in the $z$-axis direction.}
	\label{fig:ulhk}
\end{figure}

\subsection{Accuracy Experiments}

To evaluate accuracy performance, F-LOAM, KISS-ICP, and CT-ICP were selected as comparative algorithms due to their exceptional performance in the field of LiDAR odometry. The Absolute Trajectory Error (ATE) results in TABLE  \ref{tab:rmse_benchmark},  calculated using the evo \cite{evo} package, demonstrate that I$^2$EKF-LO exhibits superior accuracy across most dataset sequences. Notably, F-LOAM, lacking distortion correction for point clouds, experienced significant drift in two sequences of the NTU dataset. CT-ICP, with its continuous-time pose representation, showed consistent performance throughout the tests. Compared to the traditional Iterative Extended Kalman Filter (IEKF), which iterates only on the observation equation (results denoted as I$^2$EKF-LO(w/o i2)), I$^2$EKF-LO, through dual iteration and a process of continual prediction and distortion correction, achieves more precise point cloud data, resulting in better accuracy. The mapping results of I$^2$EKF-LO on the NTU dataset are illustrated in Fig. \ref{fig:NTU_res}. Furthermore, in the \textit{eee} and \textit{rtp} sequences of the NTU VIRAL dataset, where the sensor platform is a rotorcraft, the rotational and translational components of its state are relatively independent, showing weak coupling. Thus, the accuracy of using uniform motion model 1 outperformed that of uniform motion model 2. However, for the \textit{room} sequence of M2DGR dataset and the \textit{urbanloco\_1} sequence of Urbanloco dataset, where the sensor platform is a ground mobile robot, the rotational and translational components of its state are strongly coupled, and the results verified that uniform motion model 2 is more reasonable under these circumstances.In addition, as shown in Fig. \ref{fig:ulhk}, I$^2$EKF-LO exhibits higher $z$-axis estimation accuracy compared to the other three algorithms in the Urbanloco dataset

In the context of non-repetitive scanning solid-state LiDAR, I$^2$EKF-LO and KISS-ICP were tested on the HIT-TIB dataset, with end-to-end error serving as the evaluation metric. The results are shown in Table \ref{tab:hit-ee}, where I$^2$EKF-LO performs the best, minimizing the end-to-end error in all three sequences. In contrast, using the plain IEKF, which only iterates over the observation process, the final computed end-to-end errors are all larger than the full I$^2$EKF framework. Compared to the former two, KISS-ICP exhibits larger end-to-end errors in all three sequences. And the mapping result of \textit{walk} sequence is shown in Fig. \ref{fig:tib}, I$^2$EKF-LO performs point cloud undistortion in an iterative manner, which results in better processing of point cloud details. 

\begin{figure*}[t]
	\centering 
	\includegraphics[width=0.9\textwidth]{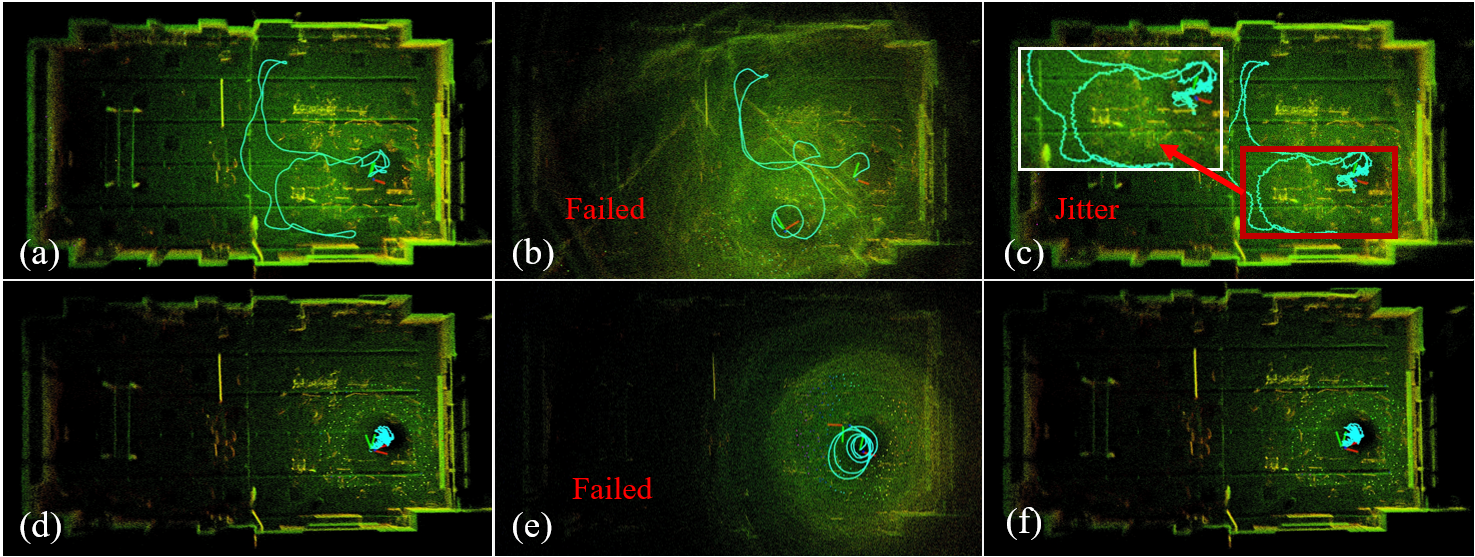}
	\caption{The mapping results for the \textit{normal} and \textit{dance} sequence are as follows: (a) - (c) correspond to the \textit{normal} sequence, where (a) shows the mapping result of the standard I$^2$EKF-LO, (b) and (c) show the mapping results using fixed process noise (0.01 and 100), respectively; (d) - (f) correspond to the \textit{dance} dataset, with (d) showing the mapping result of the standard I$^2$EKF-LO, (e) and (f) showing the mapping results using fixed process noise (0.01 and 100), respectively.}
\label{fig:akf}
\end{figure*}

\subsection{Computational Complexity Experiments}

To assess the time efficiency of I$^2$EKF-LO, tests were conducted on the mentioned datasets. The duration for each frame's state estimation was recorded under the condition of processing an equal number of points in the point cloud, with the findings summarized in TABLE \ref{tab:time}. F-LOAM, through simplification of the point cloud registration process, managed to operate at approximately 50 Hz on the \textit{room} sequence of the M2DGR dataset. Conversely, CT-ICP demonstrated the longest processing time among the compared algorithms on the same dataset sequence. I$^2$EKF-LO, utilizing a direct method without the need for feature extraction from point clouds and employing an ikd-tree within its filtering framework for map management, significantly outpaced the other reference algorithms in time efficiency when processing an equivalent number of points.

\subsection{Robustness Experiments}

To further verify the stability of I$^2$EKF-LO under various levels of motion intensity, additional datasets were recorded by hand-holding a Livox Mid-360 LiDAR, including two sequences named \textit{normal} and \textit{dance}, which correspond to regular and intense motion, respectively. As illustrated in Fig. \ref{fig:akf}, (a) and (d) display the mapping results of I$^2$EKF-LO on the \textit{normal} and \textit{dance} sequences using the same parameters, respectively. Owing to the capability to dynamically adjust process noise during state estimation, both trajectories estimated by I$^2$EKF-LO were relatively smooth, indicating high-quality mapping. (b) and (e) show the results with a fixed, smaller process noise, which failed and diverged in both the \textit{normal} (b) and \textit{dance} (e) sequences. This indicates that persistently low process noise struggles to cope with highly dynamic and intense motion scenarios. (c) and (f) present the outcomes with a larger process noise on the normal and dance sequences respectively. Although continuous high process noise prevented system divergence, the estimated trajectoriey for \textit{normal} sequences were noticeably jittery, indicating an unstable state of the system.

\section{CONCLUSION}
This paper introduces the I$^2$EKF-LO, a LiDAR odometry that employs a Dual Iterative Extended Kalman Filter. Compared with the traditional IEKF, the I$^2$EKF iterates both the observation process as well as the undistortion part of the prediction process to ensure an accurate matching relationship and a sufficient degree of linearization while adequately correcting the motion distortion in the point cloud, thus improving the accuracy of the state estimation. Additionally, we have developed a dynamic noise update module that allows the system to adapt to varying degrees of motion intensity and selects different state propagation models based on the carrier's motion pattern. Experiments conducted on public datasets and in real-world environments demonstrate that the precision and temporal efficiency of I$^2$EKF-LO are outstanding in the field of LiDAR odometry.

\section{DISSCUSSION}

The concept of dual iteration in the I$^2$EKF is equally applicable to LIO (LiDAR-Inertial Odometry) systems. The iterative updating of the accelerometer and gyroscope biases in the IMU allows for an additional iterative prediction process. This process uses the iteratively corrected biases to update the IMU's integral values, which in turn facilitates the distortion correction of the point cloud for improved data quality. This process also follows the ``chicken-laying-egg, egg-laying-chicken" logic. However, our empirical tests have shown that due to the minimal change in IMU biases within a single estimation, the impact of iterative prediction on distortion correction is not significant, leading to limited improvements in accuracy but with a high time consumption. Therefore, the I$^2$EKF is more suited to LiDAR odometry problems.

\addtolength{\textheight}{-12cm}   

\bibliography{root}

\end{document}